\documentclass{article}
\usepackage{spconf,amsmath,graphicx}
\usepackage{comment}
\usepackage[ruled, linesnumbered]{algorithm2e}
\usepackage{algorithmic}
\usepackage{amsmath,amsfonts}
\usepackage{booktabs}
\usepackage{cite}
\usepackage{multirow}
\usepackage{hyperref}
\usepackage{xurl}

\title{SRL-SOA: Self-Representation Learning with Sparse 1D-Operational \\ Autoencoder for Hyperspectral Image Band Selection}
\name{Mete Ahishali$^{\star}$ \qquad Serkan Kiranyaz$^{\dagger}$ \qquad Iftikhar Ahmad$^{\ddagger}$ \qquad Moncef Gabbouj$^{\star}$}
\address{$^{\star}$ Faculty of Information Technology and Communication Sciences, Tampere University, Tampere, Finland \\
$^{\dagger}$ Department of Electrical Engineering, Qatar University, Doha, Qatar \\
$^{\ddagger}$ Tietoevry Oy, Espoo, Finland}

\begin{document}
\ninept
\maketitle
\begin{abstract}
The band selection in the hyperspectral image (HSI) data processing is an important task considering its effect on the computational complexity and accuracy. In this work, we propose a novel framework for the band selection problem: Self-Representation Learning (SRL) with Sparse 1D-Operational Autoencoder (SOA). The proposed SLR-SOA approach introduces a novel autoencoder model, SOA, that is designed to learn a representation domain where the data are sparsely represented. Moreover, the network composes of 1D-operational layers with the non-linear neuron model. Hence, the learning capability of neurons (filters) is greatly improved with shallow architectures. Using compact architectures is especially crucial in autoencoders as they tend to overfit easily because of their identity mapping objective. Overall, we show that the proposed SRL-SOA band selection approach outperforms the competing methods over two HSI data including Indian Pines and Salinas-A considering the achieved land cover classification accuracies. The software implementation of the SRL-SOA approach is shared publicly\footnote{The software implementation of the proposed SRL-SOA approach is provided at \url{https://github.com/meteahishali/SRL-SOA}.}.
\end{abstract}
\section{Introduction}
\label{sec:intro}

Hyperspectral imaging sensors are able to capture the observed scene with hundreds of different wavelengths. Hence, these optical sensors provide rich spectral information about the target and they have been used in many applications such as target detection \cite{SpaBS2}, land-cover classification \cite{SpaBS1, EGCSR}, face recognition \cite{face}, and medical imaging \cite{medical}. However, such rich spectral information introduces several drawbacks and limitations in hyperspectral image (HSI) data processing related to computational time complexity and memory.

Due to the curse of dimensionality, it is shown in several HSI classification studies \cite{dim1, dim2} that the required number of training samples grows exponentially with the number of frequency bands. This is also called the Hughes phenomenon in HSI data \cite{hughes1968mean}. Thus, the band selection procedure plays an essential role to reduce the need for more training samples, labeling cost, and overall computational complexity in a classification framework. There have been various proposed band selection strategies in \cite{SpaBS2,SpaBS1,SpaBS3,EGCSR,ISSC}. The methods based on Self-Representation Learning (SRL) aim to represent the HSI data using the linear combination of all bands; and essentially, the obtained band coefficients in this representation will determine the importance of corresponding bands. In Sparse Representation based Band Selection (SpaBS) methods \cite{SpaBS2,SpaBS1,SpaBS3}, they set a constraint that the representation of the HSI data is sparse in the transformed domain and corresponding non-sparse coefficients denote only the most descriptive bands. In \cite{EGCSR}, Efficient Graph Convolutional Self-Representation (EGCSR) is proposed for band selection. Accordingly, they design the EGCSR model using graph convolution where the traditional model of self-representation is extended by considering each band as a node in the non-Euclidean domain over a graph. Finally, a subspace clustering method is proposed in \cite{ISSC} as the Improved Sparse Subspace Clustering (ISSC) technique. On contrary to general SpaBS methods that are based on $\ell_1$-minimizers, the ISSC method uses $\ell_2$-minimization in order to avoid too sparse solutions and consider the possible correlations within the frequency bands.

Sparse autoencoders \cite{autoencoder1,autoencoder2} have been used in different representation learning applications. These unsupervised networks learn to map the given input into a hidden representation space where the representation vector is sparse but still descriptive enough to reconstruct the original input at the output layer by the decoder part of the network. The sparsity is achieved by applying different regularization techniques such as $\ell_1$, $\ell_2$, or $k$-sparse regularizations generally on the activations of the hidden layer. However, the existing autoencoders have a common limitation: the networks are able to provide only a limited non-linear mapping due to the linear convolution operation. One can increase the network depth to learn more complex non-linear transformation functions. On the other hand, increasing the number of layers (network depth) may cause overfitting in a straightforward way considering the trivial identity mapping objective of the autoencoders.

\begin{figure*}[h]
\centering
  \includegraphics[width=0.8\linewidth]{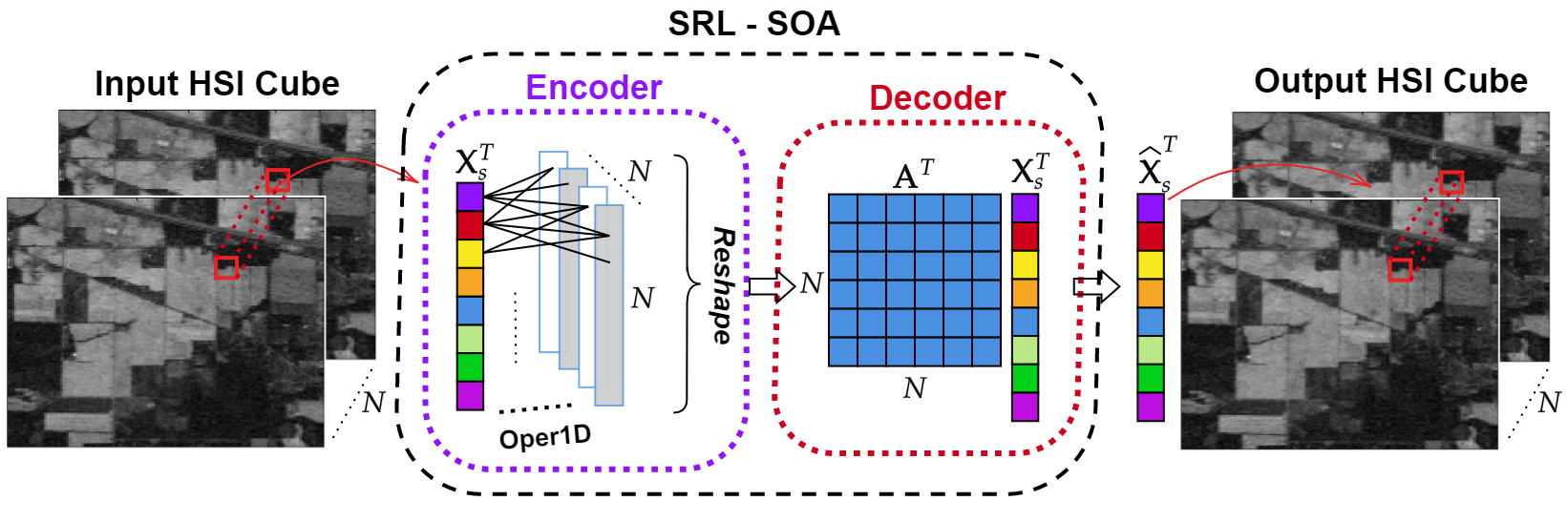}
  \vspace{-0.2cm}
  \caption{The proposed SRL-SOA framework with the 1D-operational layer where $\mathbf{X}_s$ is the batch sampled HSI cube and $\mathbf{A}$ is the learned representation matrix. The batch size is set to $m=1$ for illustration purposes. The network filters operate as expressed in \eqref{eq:operation} and it is trained with the loss function given in \eqref{cost}.}
  \label{fig:srl_soa_framework}
  \vspace{-0.2cm}
\end{figure*}

In this study, we propose a novel autoencoder model: Sparse 1D-Operational Autoencoder (SOA). Next, we design a novel band selection framework called Self-Representation Learning with Sparse 1D-Operational Autoencoder (SRL-SOA). Although the sparse autoencoders are commonly used in representation learning, they are not fully discovered for the SRL task. Hence, we show in this work that these regularized unsupervised networks can be utilized to improve the band selection performances compared to the traditional SRL approaches. Moreover, to address the abovementioned limitation of the traditional autoencoders, the proposed SOA network consists of self-organized 1D-operational layers with the generative neuron model that can perform any non-linear transformation for each kernel element. In this neuron model, the non-linear transformation function is approximated via Taylor-series expansion and the model's weights (trainable parameters) are the approximated function coefficients. It is shown that in several classification and denoising applications, the Self-Organized Operational Neural Networks (Self-ONNs) with the 2D-operational layers \cite{onn1,onn2,onn3,onn4} encapsulating generative neurons have achieved improved performance levels compared to the Convolutional Neural Networks (CNNs) with traditional linear neuron model performing only convolution. In this study, the proposed SOA is able to learn complex non-linear transformations for the SRL problem in an efficient way without requiring deep network architectures and eventually causing overfitting. The performance evaluations of the proposed SRL-SOA approach have been carried out over two HSI datasets including Indian Pines \cite{indian} and Salinas-A \cite{salinas} and it is shown that the proposed approach achieves a superior band selection performance levels compared to the SpaBS, EGCSR, ISSC, and Principal Component Analysis (PCA) approaches.

The rest of the paper is organized as follows: we first present the proposed HSI band selection approach using SRL-SOA in Section \ref{Method}. The experimental results and comparative evaluations are presented in Section \ref{Results}, and finally, Section \ref{Conclusion} concludes the paper.

\section{Proposed Methodology}
\label{Method}

In this section, we first define our objective of the sparse self-representation for the band selection problem. Then, 1D-operational layers are presented followed by the SRL-SOA model.

\subsection{Sparse Self-Representation}

Given the HSI cube $\mathbf{X} = \{ \mathbf{x}_i \}_{i=1}^N \in \mathbb{R}^{M \times N}$, consisting of $M$ number of samples and $N$ number of bands, the sparse self-representation model can be defined for $\mathbf{X}$ as follows,
\begin{equation}
\label{rep_eq}
\mathbf{X} = \mathbf{XA} + \mathbf{N},
\end{equation}
where $\mathbf{A}, \mathbf{N} \in \mathbb{R}^{N \times N}$ are the sparse representation coefficient matrix and the representation error, respectively. Accordingly, it is assumed that each band can be represented by the linear combination of other frequency bands. The sparse solution for $\mathbf{A}$ in \eqref{rep_eq} can be obtained, e.g., solving the below $\ell_0$-norm constrained problem:
\begin{equation}
\label{rep_solve}
\min_\mathbf{A} ~ \left \| \mathbf{A}\right \|_{0}~ \text{s. t.}~ \mathbf{X} \mathbf{A} + \mathbf{N} = \mathbf{X},~ \text{diag}(\mathbf{A}) = 0,
\end{equation}
where $\text{diag}(\mathbf{A}) = 0$ is used to prevent trivial solutions that each band is represented by itself. In general, the approaches \cite{EGCSR,ISSC} for SRL use the closest norm relaxation of \eqref{rep_solve} that is $\ell_1$-norm as it is more efficient and practical to solve.

\subsection{1D-Operational Layers}

A self-organized operational layer composes of generative neurons that are able to approximate non-linear kernel transformation for each 1D kernel element using Taylor series expansion. Accordingly, Taylor series can be expressed near the origin for the function $f(.)$ as,
\begin{equation}
    f(z) = \sum_{q=0}^{\infty}\frac{f^{(q)}(0)}{q!}(z)^q.
\end{equation}
Then, for $Q^{\text{th}}$ order approximation, the transformation function is as follows:
\begin{equation}
    g(x, \mathbf{w}) = w_0 + xw_1 + x^2w_2 + \dots + x^Qw_Q.
\end{equation}
where $w_q = \frac{f^{(q)}(0)}{q!}$ is the trainable parameter ($q^{th}$ coefficient of the $Q^{th}$ order polynomial).

In a 1D-operational layer, for each generative neuron, assume that $\mathbf{W}^{(k)} = [\mathbf{w}_1^{(k)}, \mathbf{w}_2^{(k)}, \dots, \mathbf{w}_Q^{(k)}] \in \mathbb{R}^{f_s \times Q}$ is the $k^{\text{th}}$ filter and $\mathbf{w}_q^{(k)} \in \mathbb{R}^{f_s}$ is the trainable parameter for $q^{\text{th}}$ coefficient with the filter size of $f_s$, then, the output of the $k^{\text{th}}$ filter is obtained by the following operation:
\begin{equation}
    \label{eq:operation}
    \mathbf{y}^{(k)} = \sigma \left( \sum_{q=1}^Q (\mathbf{x})^q * \mathbf{w}_q^{(k)} + b_q^{(k)} \right),
\end{equation}
where $*$ and $\sigma(.)$ are the 1D convolution operation and activation function (hyperbolic tangent), respectively, and $b_q$ is the bias. Overall, the $k^{th}$ neuron of the proposed layer have the following trainable parameters $\mathbf{\Theta}_k = \{\mathbf{W}^{(k)} \in \mathbb{R}^{f_s \times Q}, \mathbf{b}^{(k)} \in \mathbb{R}^{Q}\}$.

\subsection{SRL-SOA: Self-Representation Learning with Sparse 1D-Operational Autoencoder}

In the proposed SOA network, the encoder part consists of a single 1D-operational layer with L-number of filters (generative neurons) having $\mathbf{\Theta_{\text{en}}} = \{\mathbf{\Theta}_k\}_{k=1}^L$ trainable parameters. Accordingly, for a batch sampled HSI cube $\mathbf{X}_s =  [\mathbf{x_1}, \mathbf{x_2}, \dots, \mathbf{x_m}] \in \mathbb{R}^{m \times N}$ where $m$ is the batch size, the encoder provides the following mapping: $\phi\left (\mathbf{X}_s, \mathbf{\Theta_{\text{en}}} \right) = \mathbf{A}_s = [\mathbf{A}_1, \mathbf{A}_2, \dots, \mathbf{A}_m] \in \mathbb{R}^{m \times N \times N}$. In the reconstruction stage of the proposed network, we apply the self-representation pixel-wise. Specifically, the decoder part performs the following operation $\widehat{\mathbf{X}}_s = \psi(\mathbf{X}_s, \mathbf{A}_s)$ to reconstruct the batch sampled HSI cube $\mathbf{X}_s$:
\begin{equation}
\label{decoder}
    \widehat{\mathbf{X}}_s = \begin{bmatrix}\mathbf{x_1} & \mathbf{x_2} & \dots & \mathbf{x_m} \end{bmatrix} \begin{bmatrix} \mathbf{A}_1 \\ \mathbf{A}_2 \\ \vdots \\ \mathbf{A}_m\end{bmatrix}.
\end{equation}
Then, the final representation coefficient matrix is obtained by the absolute mean of the obtained multiple representation matrices over a batch: $\mathbf{A} = 1/m\sum_{i=1}^m \left| \mathbf{A}_i \right|$.

Overall, in the proposed SRL-SOA framework, the SOA model is trained to minimize the following cost using ADAM \cite{kingma2014adam} optimizer:
\begin{equation}
\label{cost}
    \begin{split}
    \mathcal{L}(\mathbf{\Theta}_{\text{en}}, \mathbf{X}_s, \widehat{\mathbf{X}}_s) = \frac{1}{2}\left \|  \mathbf{X}_s - \widehat{\mathbf{X}}_s \right \|_2^2 + \lambda \left \|\mathbf{A} \right \|_1&, \\
    ~ \text{s. t.}~ \text{diag($\mathbf{A}$)} = 0&,
    \end{split}
\end{equation}
where $\lambda$ is the regularization parameter for the trade-off between the sparsity and data fidelity parts. Note the fact that the decoder part does not have any trainable parameter as it applies the reconstruction by \eqref{decoder}. The SRL-SOA framework is illustrated in Fig. \ref{fig:srl_soa_framework}.

After the training procedure, the most informative bands are selected using only the encoder part of the network. Given the training HSI cube, $\mathbf{X}_t$, compute $\mathbf{A}_t = \phi\left (\mathbf{X}_t, \mathbf{\Theta_{\text{en}}} \right)$, then the average representation coefficient matrix is calculated over the training set: $\mathbf{A} = 1/t\sum_{i=1}^t \left| \mathbf{A}_i \right|$. Finally, the weight of $i^\text{th}$ band is obtained by $\mathbf{\alpha}_i = \sum_{j=1}^N \mathbf{A}_{i,j}$. Overall, the pseudo-code for the proposed approach is presented in Algorithm \ref{alg:selection}.

\begin{algorithm}
\caption{Band Selection with SRL-SOA.}\label{alg:selection}
\textbf{Input:} HSI data cube $\mathbf{X}$, $\lambda$, $Q$, and learning parameters. \\
\textbf{Output:} Indices of the most descriptive bands. \\
Sample the train set: $\mathbf{X}_t$, and apply normalization; \\
Initialize the trainable parameters, $\mathbf{\Theta}_{\text{en}}$; \\
\While{ $iter < $ maxIter}{
Sample the batch set $\mathbf{X}_s$ from the train set;\\
Obtain $\mathbf{A}_s = \phi(\mathbf{X}_s, \mathbf{\Theta}_{\text{en}})$ by the encoder;\\
Compute the representation matrix: $\mathbf{A} = 1/m\sum_{i=1}^m\left| \mathbf{A}_i \right|$;\\
Reconstruct: $\widehat{\mathbf{X}}_s = \psi(\mathbf{X}_s, \mathbf{A}_s)$ by the decoder in \eqref{decoder};\\
Using ADAM optimizer, calculate the updated $\mathbf{\Theta}_{\text{en}}$ minimizing the loss in \eqref{cost};\\
}
Obtain $\mathbf{A}_t = \phi(\mathbf{X}_t, \mathbf{\Theta}_{\text{en}})$ for $\mathbf{X}_t$;\\
Compute the final representation matrix $\mathbf{A} = 1/t\sum_{i=1}^t\left| \mathbf{A}_i \right|$ for $t$ number of samples;\\
Calculate the weights for each frequency band: $\{\alpha_1, \alpha_2, \dots, \alpha_N\} = \{\sum_{j=1}^N \mathbf{A}_{i,j}\}_{i=1}^N$;\\
Select the largest $k$-band indices;
\end{algorithm}

\section{Experimental Evaluation}
\label{Results}

In the experimental evaluations, two HSI datasets are used: Indian Pines \cite{indian} and Salinas-A \cite{salinas}, both are acquired by AVIRIS sensor with 224 frequency bands. The band selection performances have been evaluated based on the classification results obtained by the SVM classifier after applying the band selection procedure. In this manner, the proposed approach is compared against the following band selection methods: SpaBS \cite{SpaBS2,SpaBS1}, EGCSR \cite{EGCSR}, ISSC \cite{ISSC}, and PCA. In the following, the experimental datasets and settings are first explained and then we present the band selection performances.

\begin{figure*}[h]
\centering
  \includegraphics[width=0.9\linewidth]{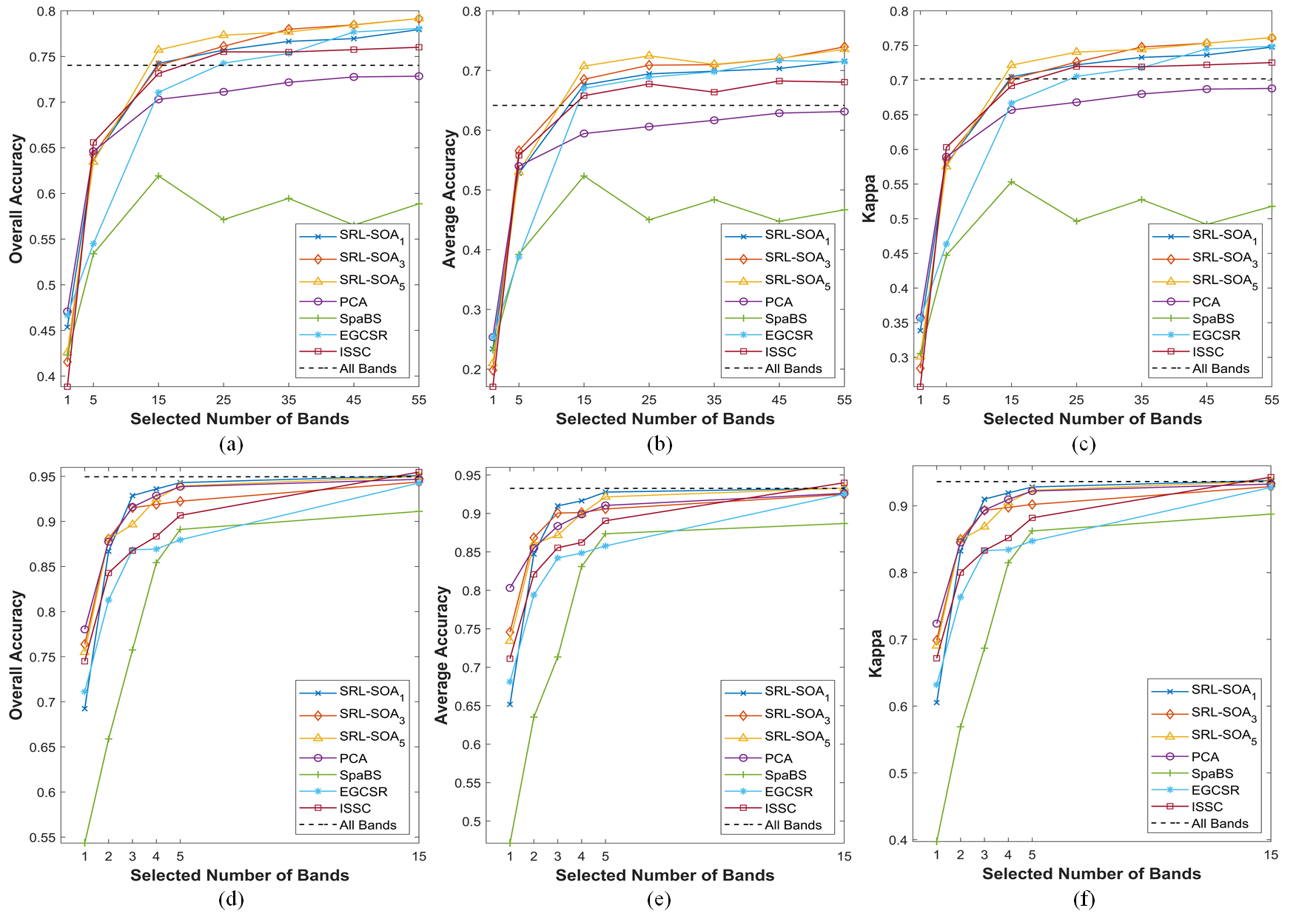}
  \vspace{-0.4cm}
  \caption{The classification results versus the selected number of bands by the proposed approach ($\text{SRL-SOA}_Q$ with $Q=1$, $3$, and $5$) and different band selection methods on the Indian Pines dataset in (a - c) and Salinas-A dataset in (d - f).}
  \vspace{-0.1cm}
  \label{fig:results}
\end{figure*}

\subsection{Datasets and Experimental Setup}

The Indian Pines scene has $145 \times 145$ pixels and 16-classes covering natural vegetation, forest, and agriculture. In the Salinas-A scene, there are $86 \times 83$ pixels and 6-classes of vegetables consisting of corn, broccoli, and lettuce in different ripeness. As followed in \cite{SpaBS3, EGCSR, ISSC}, we have also discarded the bands that cover the water absorption regions for Indian Pines and Salinas-A datasets: $\{[104-108], [150-163]\}$ and $\{[108-112], [154-167], 224\}$, reducing the number of bands to $200$ and 204, respectively. There are $10249$ annotated samples in the Indian Pines dataset and only $~5\%$ ($512$ samples) are used for training by random sampling. In the Salinas-A dataset, we randomly select only $~1\%$ ($53$ samples) of whole annotated data among annotated $5348$ number of pixels. For the Salinas-A scene, in addition to the training samples, we also use the samples that do not have any annotation during the training/fitting stages of all methods. All experiments have been repeated $10$ times including the random selection of the training samples and average performances are reported.

All the experiments have been carried out using Python and the proposed SRL-SOA approach is implemented on Tensorflow \cite{abadi2016tensorflow}. The hyperparameters of the proposed approach are set to the following values: the regularization parameter is chosen as $\lambda = 0.01$, ADAM optimizer's default parameter values are used in the training (learning rate is $10^{-3}$, $\beta_1 = 0.9$, and $\beta_2 = 0.999$) with the batch size of 5, and it is trained for 50 epochs. In the compared methods, the hyperparameter values are set to their proposed default values. The EGCSR method has two versions depending on ranking or clustering based selection over the contribution matrix. In the experiments, we use the ranking-based EGCSR as it has provided better results. The hyperparameters of the SVM classifier have been searched using the grid-search technique in each individual run with performing 2-fold cross-validation over the training set. In the grid-search, the following set of kernel functions and parameters are included: the SVM decision scheme: one-versus-one and one-versus-all, kernel function \{Linear, Radial Basis Function (RBF), Polynomial\}, the box constraint $C$ parameter in the range of $[10^{-3}, 10^{-3}]$ incremented in log-scale, the $\gamma$ parameter of the RBF kernel $[10^{-3}, 10^{-3}]$ incremented in log-scale, and the polynomial degree in $\{2, 3, 4\}$.

\subsection{Results}

In the proposed SRL-SOA approach, we choose the degree of the polynomial as $Q=1$, $3$ and $5$ ($\text{SRL-SOA}_1$, $\text{SRL-SOA}_3$, and $\text{SRL-SOA}_5$) and compare the classification results with different band selection approaches in Fig. \ref{fig:results}. It is observed that the proposed approach significantly outperforms all competing methods. For example, the best average accuracy (AA) in the Indian Pines dataset is obtained by $\text{SRL-SOA}_3$ at 55 bands as $73.96\%$ which is $9 - 9.5\%$ higher than the AA obtained by using \textit{all} bands. Compared with the other band selection methods, $\text{SRL-SOA}_3$ produces $5.5\%$ larger than the best AA achieved by the best performing competing method ISSC on the Indian Pines dataset. The improved classification results by $\text{SRL-SOA}_5$ is especially noticeable between the bands $15$ and $35$ in the Indian Pines dataset. On the other hand, the classification problem is more accessible on the Salinas-A dataset than the Indian Pines as there are only 6-classes. Hence, using 15 or 25 bands produce comparable results for all methods. In this case, classification results are provided for the selected number of bands less than $5$ to provide a better comparison between the competing methods. It is observed that using less than $5$ bands in the Salinas-A dataset, only the proposed approach is able to provide a classification accuracy greater than $90\%$ except for the PCA method. However, one can say that the PCA method is actually a feature extraction technique rather than a feature selection method. In SRL-SOA, SpaBS, EGCSR, and ISSC approaches, we first apply band selection on the training set and decide the most descriptive band labels. Then, we simply choose the same bands on the test data for the classification. Therefore, there is no inference time on the test set (only indexing operation). On the other hand, the inference stage of the PCA method requires a significant computation time and memory as the PCA matrix is computed on the train set, and only then it is applied to the test data.

\begin{table}[h!]
\vspace{-0.2cm}
\caption{Classification results: overall accuracy (OA), average accuracy (AA), and Kappa values for the proposed $\text{SRL-SOA}_Q$ approach with $Q=1,3,5$ and competing methods on the Indian Pines ($25$ bands selected) and Salinas-A ($2$ bands selected).}
\label{tab:results}
\resizebox{\linewidth}{!}{
\begin{tabular}{@{}c|ccc|ccc@{}}
\toprule
\textbf{}       & \multicolumn{3}{c|}{\textbf{Indian Pines}}          & \multicolumn{3}{c}{\textbf{Salinas-A}}              \\ \midrule
\textbf{Method} & \textbf{OA}     & \textbf{AA}     & \textbf{Kappa}  & \textbf{OA}     & \textbf{AA}     & \textbf{Kappa}  \\ \midrule
$\text{\textbf{SRL-SOA}}_\mathbf{1}$           & 0.7570          & 0.6944          & 0.7220          & 0.8670          & 0.8473          & 0.8323          \\
$\text{\textbf{SRL-SOA}}_\mathbf{3}$           & 0.7612          & 0.7090          & 0.7261          & \textbf{0.8807} & \textbf{0.8684} & \textbf{0.8499} \\
$\text{\textbf{SRL-SOA}}_\mathbf{5}$           & \textbf{0.7733} & \textbf{0.7247} & \textbf{0.7404} & 0.8802          & 0.8603          & 0.8486          \\
PCA             & 0.7113          & 0.6061          & 0.6680          & 0.8775          & 0.8549          & 0.8453          \\
SpaBS           & 0.5714          & 0.4501          & 0.4963          & 0.6588          & 0.6352          & 0.5691          \\
EGCSR           & 0.7426          & 0.6887          & 0.7055          & 0.8128          & 0.7941          & 0.7632          \\
ISSC            & 0.7551          & 0.6775          & 0.7197          & 0.8429          & 0.8208          & 0.8002          \\ \midrule
All Bands       & 0.7403          & 0.6416          & 0.7018          & 0.9495          & 0.9326          & 0.9363          \\ \bottomrule
\end{tabular}
}
\end{table}

In Table \ref{tab:results}, the classification results are given when $25$ bands are chosen for the Indian Pines and $2$ bands for the Salinas-A. For the proposed approach with 25 bands, it is observed that increasing the $Q$ value improves the performance in the Indian Pines and the best results are obtained by $\text{SRL-SOA}_5$. On the other hand, in the Salinas-A dataset with 2 bands, $\text{SRL-SOA}_3$ gives the best classification result. Note that when fewer bands are used for the classification, the band selection task becomes more challenging and the performance gaps between the proposed approach and competing methods are larger in the Salinas-A dataset. Among them, only the PCA method can produce comparable results in the Salinas-A dataset considering OA and Kappa performance metrics. Recalling the fact that PCA is used for the feature extraction, whereas we design the band selection problem as a feature selection method; the proposed approach enjoys the improved performance and no-inference time on the test set.

\section{Conclusion}
\label{Conclusion}

Band selection task plays an essential role in HSI data processing with scarce data. In this work, we propose a novel band selection approach for HSI images: SRL-SOA. The proposed approach consists of a novel sparse autoencoder model, SOA, that is designed for the SRL problem. Thanks to the operational layer in the SOA, the improved neuron models can efficiently learn non-linear kernel transformation functions and provide better SRL delivering the advanced band selection performance. The experimental evaluations have been performed on the Indian Pines and Salinas-A HSI datasets when $5\%$ and $1\%$ training data is used in the classification, respectively. The proposed SRL-SOA approach outperforms all competing band selection methods in all performance metrics used.


\vfill\pagebreak

\bibliographystyle{IEEEbib}
\bibliography{refs}

\end{document}